\documentclass[letterpaper]{article} 
\usepackage[]{aaai2026}  
\usepackage{times}  
\usepackage{helvet}  
\usepackage{courier}  
\usepackage[hyphens]{url}  
\usepackage{graphicx} 
\usepackage{natbib}  
\usepackage{caption} 
\usepackage{algorithm}
\usepackage{algorithmic}

\usepackage{newfloat}
\usepackage{listings}
\DeclareCaptionStyle{ruled}{labelfont=normalfont,labelsep=colon,strut=off} 
\floatstyle{ruled}
\newfloat{listing}{tb}{lst}{}
\floatname{listing}{Listing}
\nocopyright

\usepackage{mathtools}
\usepackage{amssymb}
\usepackage{booktabs}
\usepackage{subfigure}

\usepackage{xcolor}

\title{Inverse Manipulation through Symbolic Planning and Residual Operator Learning}
\author {
    Yigit Yildirim\textsuperscript{\rm 1},
    Giuseppe Rauso\textsuperscript{\rm 2},
    Riccardo Caccavale\textsuperscript{\rm 1, 2},
    Alberto Finzi\textsuperscript{\rm 1, 2}
}
\affiliations {
    \textsuperscript{\rm 1} CREATE Consortium, Italy\\
    \textsuperscript{\rm 2}Department of Electrical Engineering and Information Technologies,
Università di Napoli Federico II, Italy\\
    yigit.yildirim@consorziocreate.it, \{giuseppe.rauso,riccardo.caccavale,alberto.finzi\}@unina.it
}

\begin{document}

\maketitle

\begin{abstract}

Inverting a robotic task requires more than reversing symbolic state transitions or rewinding motor trajectories. In robot manipulation tasks, symbolic inverse plans often fail to fully restore the effects of forward executions under continuous interaction dynamics. We present a hybrid framework for inverse manipulation that derives inverse-skill objectives from STRIPS-like operators automatically extracted from demonstrations through soft geometric predicates. For each extracted operator, we construct an inverse restoration objective that preserves preconditions, restores delete effects, and negates add effects. A task planner first attempts to satisfy this objective using available action primitives. Unresolved symbolic predicates then induce a residual operator learning problem solved through Reinforcement Learning (RL). We evaluate the framework on the ManiSkill3 PushCube task. For a forward pushing skill, the symbolic inverse performs a coarse pick-and-place restoration, while a residual Soft Actor-Critic policy refines the cube pose to satisfy the remaining inverse predicates. Our results show that predicate-derived residual control can turn an approximate symbolic inverse into a physically grounded inverse skill.

\end{abstract}


\section{Introduction}
Human-robot collaboration and robotic assembly-disassembly lines share a common requirement: the ability to undo previously executed skills. However, undoing a task can go well beyond simply rewinding the forward skill's motor trajectory, which is a viable option for only a subset of inversion problems \cite{eysenbach2018leave}. A more generic definition of undoing is restoring the world to its pre-forward state.

State restoration can be achieved in different ways. Consider a pick-and-place skill in which the robot picks an object from a position, carries it, and drops it at another location. The inverse of a pick-and-place skill 
is another pick-and-place skill, 
which can often be achieved through a symmetric manipulation strategy.
Nevertheless, inverting some other skills requires more than simply rewinding the execution \cite{grinsztajn2021there}. For instance, a non-prehensile push cannot be inverted by retracting the manipulator to its original position, because the robot does not grasp the object, \cite{mason2018toward}. This is an inter-class inversion that requires restoring the world to its state before executing the forward skill. From this perspective, the difference between trajectory-level and task-level inversion is more pronounced.
Task-level inversion is more challenging because it requires an additional planning process, along with physically feasible continuous manipulation, to restore the world to its original state and undo the effects of forward execution.


In this work, we propose a framework to address both in-class and inter-class inversion problems and report empirical results. 
Inspired by \cite{silver2022learning}, we first extract the forward skill from demonstrations as a symbolic operator in the STRIPS formalism \cite{fikes1971strips} using a predefined predicate repertoire.
The symbolic difference between post and preconditions becomes the restoration objective for the inverse task. 
Classical planning
is then used together with scripted action primitives to generate an action sequence that restores the objective supported by the available operator repertoire.
Since the symbolic action space is limited, the system occasionally fails to successfully undo forward executions. In such cases, the system evaluates the state in which a portion of the restoration objective is satisfied. The rest constitutes the residual objective, which is learned by Reinforcement Learning (RL). Consequently, the inverse operator is obtained by applying the action sequence returned by the planner, followed by the residual policy learned by the RL. The RL phase attempts to recover the residual conditions while keeping the already satisfied ones.

The approach is evaluated on a scripted scenario and a simulated benchmark, named ManiSkill3 \cite{tao2025maniskill3}. The preliminary results show that it can successfully plan and apply the inverse operator that requires contact-rich manipulation of an object, achieving 90\% success at a 1 cm tolerance with a 1.4 mm mean precision; whereas the symbolic prefix alone leaves a 17 mm residual, and a pure trajectory-inverse baseline is inapplicable due to a non-prehensile forward characteristic.

The contribution of this study is threefold:
\begin{itemize}
    \item We address skill inversion as a hybrid symbolic-continuous restoration problem by deriving inverse targets 
    from symbolic forward operators extracted from demonstrations.
    \item 
    We introduce a residual operator learning framework in which RL is activated only when symbolic planning cannot fully satisfy the inverse objective.
    \item  
    We derive RL rewards for residual operator learning from the symbolic plan.
\end{itemize}

\section{Related Work}
Skill inversion has been addressed in the literature through various approaches. From the symbolic planning perspective, action reversibility has been extensively studied~\cite{Eiter07, Morak20, Chrpa21}. In this work, we address the problem of reversing multi-step manipulation tasks executed by a robotic system, where incomplete reversibility is handled through residual operator learning.

 \cite{nair2020trass} uses self-supervision to learn time-reversal models from forward demonstrations to bootstrap forward policy learning. Similarly, \cite{edwards2018forward} proposes Forward-Backward RL to leverage imagined reversal steps to accelerate the forward policy modeling. They rely on temporal reversion as a self-supervision signal. \cite{hoffman1989automated, tian2022assemble, widulle2023using} leverage physical or geometric coupling between assembly and disassembly. These are successful approaches, but their assumption of time reversal limits their applicability to low-level skill inversion. In contrast, we treat inverse skills as distinct symbolic operators with their own restoration objectives. We suggest that inverse skills can be derived from the effects of forward demonstrations without relying on temporal or physical reversal.

To model symbolic operators directly from skill demonstrations, a detailed approach is presented in \cite{konidaris2018skills}. \cite{cresswell2013acquiring} proposes another system, named Learning Object Centred Models, for generating planning-domain models from demonstrations. Similarly, \cite{aineto2019learning} proposes FAMA, a novel system to learn STRIPS models from skill executions. These studies do not address skill inversion as a derived planning problem.

Even with the appropriate modeling of the symbolic operators and suitable planning, the robot may still fail to fully undo a forward execution due to missing skills or precision requirements. Addressing the residual part requires translating the unsatisfied symbolic restoration objectives into continuous RL reward signals. Recent neurosymbolic studies \cite{icarte2022reward, camacho2019ltl, de2020restraining} compile task specifications into rewards. They derive the reward from the agent's goal. We follow this thread but derive the reward from operator effects rather than task specifications, with the inverse target carrying both the goal and the constraint set.

\section{Method}

\begin{figure}[t]
    \centering
        \includegraphics[width=\columnwidth]{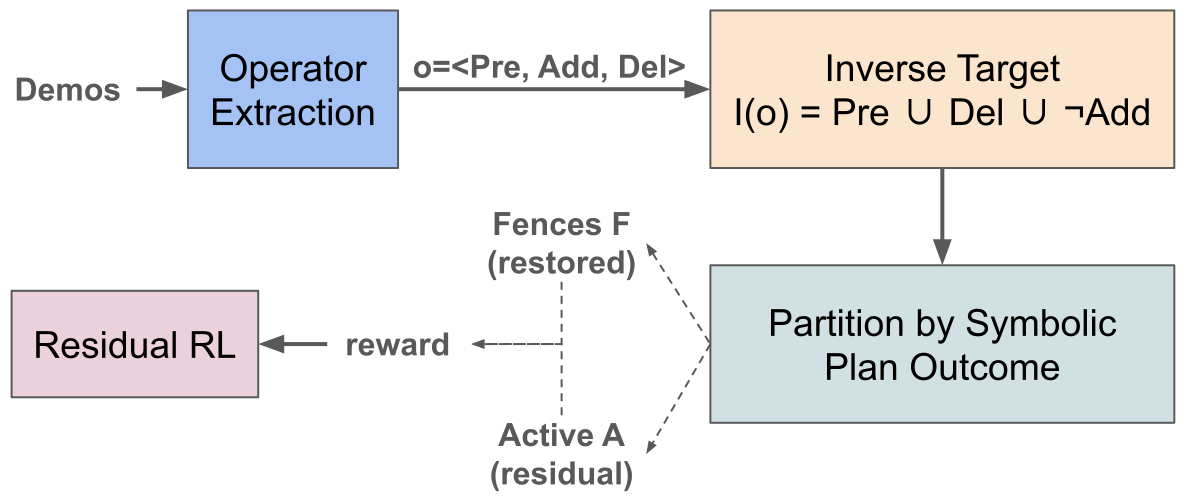}
        \caption{The overview of the proposed system. Skill demonstrations are used to extract symbolic operators. After the inverse target for the operator is determined, the components of the reward are calculated. This reward is optimized using RL.}
    \label{fig:overview}
\end{figure}

The proposed system, depicted in Figure \ref{fig:overview}, starts by creating a scene graph, in which the state of each object and the robot is recorded as continuous values. It uses soft predicates to assess the scene graph in a continuous and differentiable manner. The predefined repertoire of predicates is grounded given the scene graph. The grounding of a predicate $p$ returns a soft score, $V_p(s) \in [0, 1]$. This value is computed as in Equation \ref{eq:soft_score}, using a sigmoid with a signed margin $m_p(s) \in \mathbb{R}$, which measures how far the current state is from satisfying the predicate, and a predefined temperature.

\begin{equation}
    V_p(s) = \sigma\left(\frac{m_p(s)}{T_p}\right) = \frac{1}{1 + \exp(-m_p(s)/T_p)}
    \label{eq:soft_score}
\end{equation}

An operator is defined as a tuple, \(o = \langle Pre,\ Add,\ Del \rangle\), where \(Pre\), \(Add\), and \(Del\) denote preconditions, add list, and delete list, respectively. Operator extraction describes the process of modeling these components based on the physical execution of robotic skills. In our model, this process can use one or more demonstrations to evaluate the initial and final scenes of a skill, where more demonstrations increase its reliability. Given a list of initial and final scenes of a robotic skill, each predicate $p$ is evaluated to obtain $V_p(s)$ values to denote corresponding start and end values. These scores are averaged across demonstrations for each predicate to determine if a predicate belongs to the \(Pre\) set. Also, mean end scores are used to understand the delta between the start and the end scenes, which, in turn, are used to decide if a predicate belongs to \(Add\) or \(Del\) sets. After the extraction process is completed, the symbolic operator is stored inside the system.

Correspondingly, we define the inverse target as a restoration problem. The aim of the inverse operator to restore the world to the forward skill's preconditions. Therefore, the inverse target is constructed as in Equation \ref{eq:inverse_target}, where the preconditions \(Pre\) should be satisfied for the world to enable the applicability of the forward operator. The delete effects \(Del\) must be restored, and the add effects \(Add\) must be negated \(\neg Add\).

\begin{equation}
    \mathcal{I}(o) = Pre \cup Del \cup \neg Add
    \label{eq:inverse_target}
\end{equation}

Given the inverse target and a limited set of scripted action primitives, a simple BFS-based planner is employed to generate a plan. 
Nevertheless, given the limited set of known actions, only some predicates in $\mathcal{I}(o)$ are restored by the generated plan; some inverse targets cannot be fully met. 
In such cases, the proposed system ends with a partial solution 
leading to a handoff state for residual skill learning.

In other words, the symbolic plan
is 
an executable sequence of scripted action primitives
$\pi_{\text{sym}}$ that, given the current state $s_0$, returns a state $s_h$ (referred to as the handoff). 
Relative to $s_h$, predicates already restored by symbolic planning become fence conditions to preserve, while unresolved predicates define the active residual objective:

\begin{itemize}
    \item Fence set: $\mathcal{F} = { p \in \mathcal{I}(o): V_p(s_h) \geq \theta }$,
    \item Active residual set: $\mathcal{A} = \mathcal{I}(o) \setminus \mathcal{F}$.
\end{itemize}

where $\theta \in [0, 1]$ is a hyperparameter, and predicates in $\mathcal{F}$ can be restored by the symbolic plan, 
$\pi_{\text{sym}}$, while predicates in $\mathcal{A}$ are left for RL. A handoff is valid if every predicate in a chosen subset $\mathcal{C} \subseteq \mathcal{I}(o)$ satisfies a tolerance check: e.g., $|m_p(s_h)| < \epsilon_p$.
If the handoff is invalid, $\pi_{\text{sym}}$ is re-executed with a new seed. Therefore, the RL phase only sees valid handoffs.

For each term $(p, \sigma_p, c_p) \in \mathcal{I}(o)$, where $\sigma_p \in \{+1, -1\}$ and $c_p$ respectively denotes the sign and scale of the predicate $p$, we define a signed normalized margin
as in Equation \ref{eq:signed_normalized_margin}. The function $tanh$ provides bounded rewards so the value function does not diverge during exploration.

\begin{equation}
    \tilde m_p(s) = \sigma_p \cdot \tanh\left(\frac{m_p(s)}{c_p}\right) \in [-1, +1]
    \label{eq:signed_normalized_margin}
\end{equation}

\begin{figure}[b]
    \centering
        \includegraphics[width=\columnwidth]{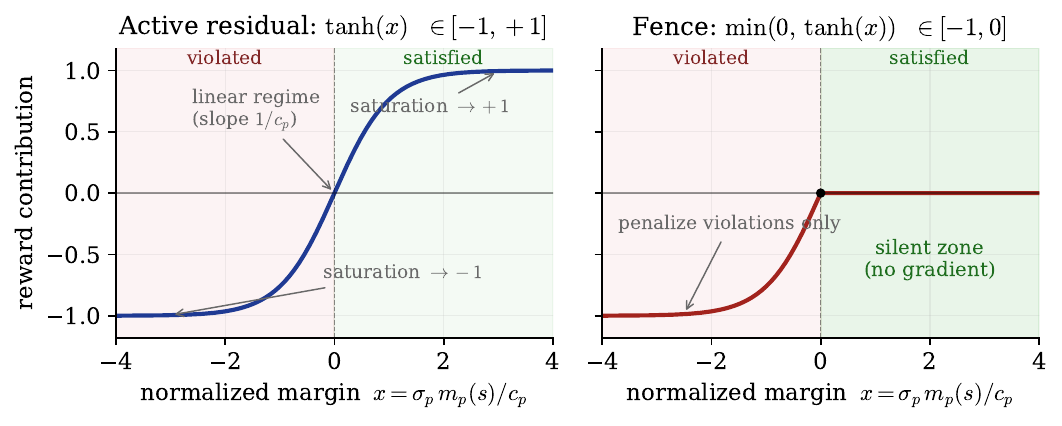}
        \caption{The reward for the residual RL. The contribution of the active predicates is computed by tanh() while the violation of the satisfied predicates is penalized.}
    \label{fig:reward_shape}
\end{figure}

Subsequently, the residual reward is defined as in Equation \ref{eq:residual_reward}. It comprises two parts, as depicted in Figure \ref{fig:reward_shape}, to stimulate the restoration of the predicates in $\mathcal{A}$, while $min(0, .)$ is used for penalizing the violation of the fences in $\mathcal{F}$.

\begin{equation}
    R(s) = \underbrace{\sum_{p \in \mathcal{A}} \tilde m_p(s)}_{\text{active},\in[-1,+1]} + \underbrace{\sum_{p \in \mathcal{F}} \min\bigl(0, \tilde m_p(s)\bigr)}_{\text{fence (one-sided)},\in[-1,0]}
    \label{eq:residual_reward}
\end{equation}

With the predicate-grounded reward and active/fence partition fixed once, the RL problem reduces to a standard continuous control problem. Note that $\mathcal{A}$ and $\mathcal{F}$ sets are populated automatically at run time, eliminating the need for per-task reward engineering. Consequently, the inverse skill is obtained by the concatenation of $\pi_{\text{sym}}$ and $\pi_{\text{res}}$.

\subsection{Implementation Details}
The system uses Stable-Baselines3 \cite{raffin2021stable} implementation of the Soft Actor-Critic, an off-policy deep RL algorithm \cite{haarnoja2018soft}. For the simulation tests, 12-dimensional, high-level observations are used, containing a predicate-grounded feature vector. 4-dimensional, continuous actions denote XYZ delta for the end effector and the gripper width, without per-task action restrictions. To promote convergence and generalizability, a curriculum during SAC training is used to control the handoff perturbation in the schedule.

\section{Experiments and Results}
We evaluate the system on the PushCube task of the ManiSkill3 framework with a Franka Panda manipulator under pd-ee-delta-pos control. The forward skill, push(cube), is a scripted routine that moves a cube on the table in the +x dimension until it reaches the goal position. Given the predefined predicate repertoire, the system is provided with different forward skill executions, and the forward operator is automatically extracted.

For the PushCube task, we use the following predicates:
\begin{align*}
  &\textsc{at\_pose}(o, p): m = \tau - \|x_o - p\| \\
  &\textsc{gripper\_open}(): m = w - w_{\min} \\
  &\textsc{tcp\_near}(o): m = \delta - \|x_{tcp} - x_o\|
\end{align*}
where $\tau, w_{\min}, \delta$ are predicate-specific thresholds.

\paragraph{Extracted forward operator.}
Given three demonstrations, the extracted operator for the forward \textsc{PUSH} skill in STRIPS form is as follows:
\begin{equation*}
\begin{aligned}
&\textsc{PUSH}(\mathit{cube}, \mathit{src}, \mathit{goal}) = \\
&\quad - \textbf{Pre: } \textsc{AT\_POSE}(\mathit{cube}, \mathit{src}) \land \\ 
&\qquad \textsc{TCP\_NEAR}(\mathit{cube}) \land \textsc{GRIPPER\_OPEN}() \\
&\quad - \textbf{Add: } \textsc{AT\_POSE}(\mathit{cube}, \mathit{goal}) \\
&\quad - \textbf{Del: } \textsc{AT\_POSE}(\mathit{cube}, \mathit{src})
\end{aligned}
\end{equation*}

\paragraph{Derived inverse target.}
Applying the inverse target generation, $\mathcal{I}(o) = \mathrm{Pre}(o) \cup \mathrm{Del}(o) \cup \neg\mathrm{Add}(o)$ to $\textsc{push}$ returns four restoration predicates:
\begin{align*}
\mathcal{I}(\textsc{push}) = \{
  &\textsc{at\_pose}(\mathit{cube}, \mathit{src}),
   \textsc{tcp\_near}(\mathit{cube}), \\
  &\textsc{gripper\_open}(),
   \neg\textsc{at\_pose}(\mathit{cube}, \mathit{goal})\}.
\end{align*}

Using scripted \textsc{pick}($\mathit{cube}$) and \textsc{place}($\mathit{src}$) primitives, the planner executes and finds out that three out of four restoration predicates can be satisfied, leaving \textsc{at\_pose}($\mathit{cube}$, $\mathit{src}$) as the active residual for the RL.

\begin{table}[h]
\begin{tabular}{@{}lll@{}}
\toprule
\multicolumn{1}{c}{Method}    & \multicolumn{1}{c}{\begin{tabular}[c]{@{}c@{}}Distance  \\ (mm, mean$\pm$std)\end{tabular}} & \multicolumn{1}{c}{\begin{tabular}[c]{@{}c@{}}Success \\ @ 1 cm\end{tabular}} \\ \midrule
Symbolic prefix (no RL)  & 16.6 $\pm$ 6.6                                & 10\%                               \\
Symbolic + random action   & 18.8 $\pm$ 17.2                               & 50\%                               \\
\textbf{Symbolic + RL (ours)} & \textbf{1.4 $\pm$ 3.2}                        & \textbf{90\%}                      \\ \bottomrule
\end{tabular}
\caption{Results with 10 different seeds.}
\label{tab:headline}
\end{table}

\begin{figure}[t]
    \centering
    \subfigure[]{\includegraphics[width=0.23\textwidth]{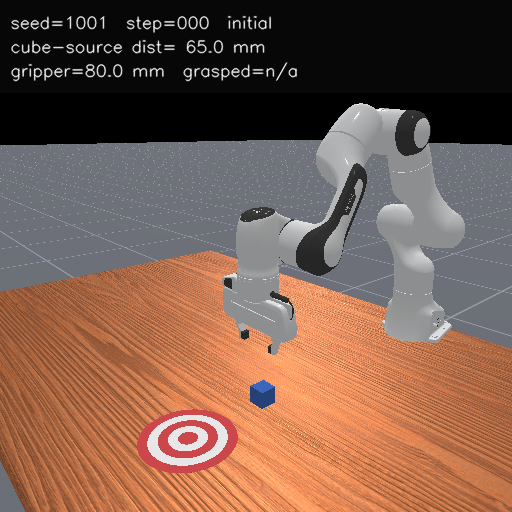}} 
    \subfigure[]{\includegraphics[width=0.23\textwidth]{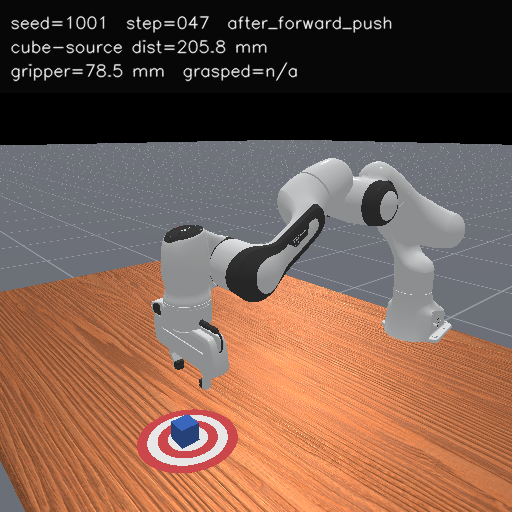}} 
    \subfigure[]{\includegraphics[width=0.23\textwidth]{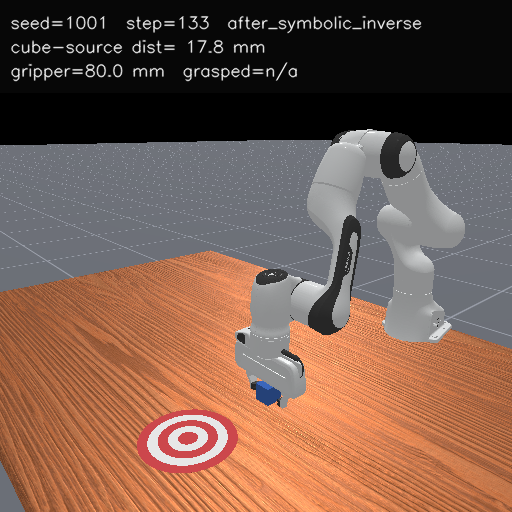}}
    \subfigure[]{\includegraphics[width=0.23\textwidth]{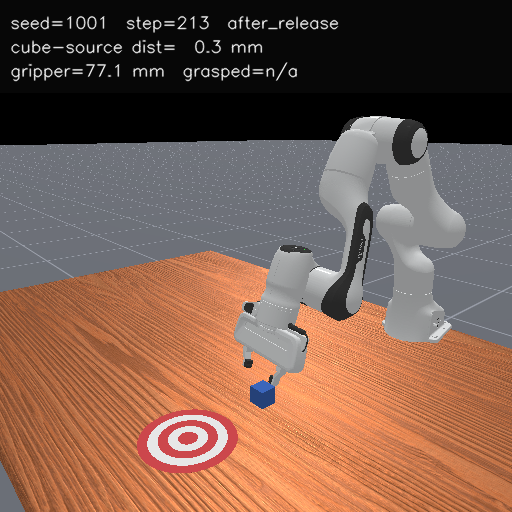}}
    \caption{Screenshot from a single evaluation: (a) the cube sits at the initial position, (b) after the forward execution, (c) after the symbolic prefix, (d) after the RL policy is applied.}
    \label{fig:run}
\end{figure}

We report in Table \ref{tab:headline} the mean $\pm$ standard deviation (std) across 10 evaluation seeds with the $\pm$2 cm handoff perturbation enabled. These results show that our method achieves very precise placement with a 90\% success rate under the 1 cm threshold of the \textsc{at\_pose}($\mathit{cube}$, $\mathit{src}$) predicate, whereas random control occasionally reaches the target configuration by chance with the symbolic phase already placing the cube within approximately 17 mm of the source. This coincidental success is further supported by the high standard deviation value. Screenshots from a successful run are given in Figure \ref{fig:run}.

\subsection{Ablations}
We report the success rate and failure modes in Table \ref{tab:ablation} across different RL reward settings to motivate our reward design. Using the regular tanh function for the fence predicates rewards the do-nothing behavior. Completely removing tanh leads to unsaturation and makes the active term unbounded, resulting in large reward drifts that the SAC policy cannot recover from. Using one-sided tanh for the fences provided the necessary and sufficient penalty to encourage exploration of a policy that satisfies the active predicate without violating the fences.

\begin{table}[h]
\begin{tabular}{@{}lll@{}}
\toprule
\multicolumn{1}{c}{Reward}                                                                          & \multicolumn{1}{c}{\begin{tabular}[c]{@{}c@{}}Success \\ @ 1 cm\end{tabular}} & \multicolumn{1}{c}{Failure Mode}                                                             \\ \midrule
Fences (with tanh)                                                                                  & 10\%                                                                          & \begin{tabular}[c]{@{}l@{}}Saturated fences reward\\ the agent for staying still\end{tabular}         \\
\begin{tabular}[c]{@{}l@{}}Unbounded margin\\ (no tanh)\end{tabular}                                & 0\%                                                                           & \begin{tabular}[c]{@{}l@{}}Value function explodes;\\ cube ends \textgreater{}100 mm off\end{tabular} \\
\textit{\textbf{\begin{tabular}[c]{@{}l@{}}Ours (active+fences \\ with min(0, tanh))\end{tabular}}} & \textbf{90\%}                                                                 & \textbf{-}                                                                                            \\ \bottomrule
\end{tabular}
\caption{Ablations with 10 different seeds.}
\label{tab:ablation}
\end{table}

\section{Conclusion}
In this paper, we presented a hybrid system that integrates symbolic planning and deep RL to address the skill inversion problem. The system automatically extracts symbolic operators from forward skill demonstrations and generates the inverse target. After applying the inverse plan, the system assigns unsatisfied predicates to a residual RL loop using a generic, predicate-based reward. The system is empirically validated on a single benchmark, and the results are reported. Specifically, the proposed system achieved a 90\% success rate at 1 cm precision. The limited baseline comparison hints at this system's promise. Therefore, results imply that an inverse skill can be solved as a planning problem followed by a residual RL problem with a derived reward.

\subsection{Limitations and Future Work}
The presented approach is currently being developed. Therefore, it has some limitations that we plan to address in the future. First, it relies on hand-designed predicates. Margins and scales are specified and not learned from observation. Predicates and related parameters can be learned from demonstrations using state-of-the-art approaches, such as \cite{athalye2026pixels}.

Currently, the system relies 
on scripted low-level action primitives for execution. In particular, the pick-and-place primitive is task-specific. In future work,
we plan to model primitives using learning-from-demonstration approaches, such as \cite{yildirim2024conditional}.
In addition, the system is evaluated only on one task family. 

\bibliography{aaai2026}

\end{document}